\documentclass[conference, a4paper]{IEEEtran}
\IEEEoverridecommandlockouts
\usepackage{cite}
\usepackage{amsmath,amssymb,amsfonts}
\usepackage{algorithmic}
\usepackage{graphicx}
\usepackage{textcomp}
\usepackage{xcolor}
\usepackage{caption}
\usepackage{subcaption}
\usepackage{optidef}
\def\BibTeX{{\rm B\kern-.05em{\sc i\kern-.025em b}\kern-.08em
    T\kern-.1667em\lower.7ex\hbox{E}\kern-.125emX}}

\begin{document}

\title{Towards Explaining Adversarial Examples Phenomenon in Artificial Neural Networks
\thanks{Preprint submitted to \textit{ICPR 2020}.}}

\author{\IEEEauthorblockN{1\textsuperscript{st} Ramin Barati}
\IEEEauthorblockA{\textit{Computer engineering Department} \\
\textit{Amirkabir university of Technology}\\
Tehran, Iran \\
ramin.barati@aut.ac.ir}
\and
\IEEEauthorblockN{2\textsuperscript{nd} Reza Safabakhsh}
\IEEEauthorblockA{\textit{Computer engineering Department} \\
\textit{Amirkabir university of Technology}\\
Tehran, Iran \\
safa@aut.ac.ir}
\and
\IEEEauthorblockN{3\textsuperscript{rd} Mohammad Rahmati}
\IEEEauthorblockA{\textit{Computer engineering Department} \\
\textit{Amirkabir university of Technology}\\
Tehran, Iran \\
rahmati@aut.ac.ir}
}

\maketitle

\begin{abstract}
In this paper, we study the adversarial examples existence and adversarial training from the standpoint of convergence and provide evidence that pointwise convergence in ANNs can explain these observations. The main contribution of our proposal is that it relates the objective of the evasion attacks and adversarial training with concepts already defined in learning theory. Also, we extend and unify some of the other proposals in the literature and provide alternative explanations on the observations made in those proposals. Through different experiments, we demonstrate that the framework is valuable in the study of the phenomenon and is applicable to real-world problems.
\end{abstract}


\section{Introduction}
With the recent advancements in processing power and algorithms, machine learning has become one of the more popular tools in many industries. It is arguable that most of this attention is due to the combined effect of deep learning and big training sets. The effectiveness of applying deep learning to common machine learning tasks like image classification, object recognition, natural language processing and speech recognition has fortified the position of multi-layer perceptron (MLP) among the standard tools of solving machine learning problems.

Considering these achievements, it would be natural to use artificial neural networks (ANNs) in applications like autonomous driving where safety and security is of importance. However, it was shown by Szegedy et al. that deep networks are sensitive to adversarial examples\cite{szegedy2013intriguing}. Adversarial examples are points close to a natural example in which the classifier output changes drastically. In some cases, the difference between the adversarial example and the natural example is not perceivable to humans.

In the near future, it is expected that many instances of MLPs are embedded in the physical world. However, research has shown that adversarial samples can be misused in physical world as well\cite{DBLP:journals/corr/KurakinGB16}. For example, the input signal can be changed in a way that a speech recognition system order a wrong action\cite{Schoenherr2019} or attack a self-driving car autonomous driving control\cite{DBLP:journals/corr/abs-1903-05157}. As a result, many concerns has been raised in regards to the use of ANNs in safety-critical applications\cite{Yuan_2019}.

In practice, ANNs are considered black-box models. Although these models work well in many tasks, they have proved to be difficult to explain and interpret. Providing an explanation for the existence of adversarial examples is crucial in minimizing their effect. But, a combination of difficulties in explaining the inner workings of ANNs and visualization of high dimensional spaces has made the problem hard to approach. Explaining the existence of adversarial examples is still an open question and we refer the reader to \cite{Yuan_2019} for a more comprehensive study of research done on other aspects of this phenomenon.

In this paper, we describe a new perspective on the reasons behind the existence of adversarial examples and discuss the consequences of our approach. This paper is organized in five sections. After introduction, in Section \ref{sec:related}, we provide a literature study of the relevant proposed explanations of this phenomenon. Next, in Section \ref{sec:sing}, we will describe the proposed perspective and will continue by providing evidence in support of the proposal in Section \ref{sec:exp}. We will finish the paper with a discussion of the consequences and possible next steps in Section \ref{sec:diss}.
%------------------------------------------------

\section{Related work}
\label{sec:related}
There have been multiple proposals for explaining the existence of adversarial examples phenomenon. The first explanation was proposed in \cite{szegedy2013intriguing}. Szegedy et al. viewed the adversarial examples as pockets in the input space which have a low probability of being observed and correctly classified. This view has been further described with an analogy between the real numbers as the natural samples and the rational numbers as the adversarial examples. In other words, There is a rational number in the neighborhood of every real number even though the probability of observing a rational number in an interval of the real line is zero. Nevertheless, as mentioned in \cite{tanay2016boundary}, the argument does not justify why would a classifier show such a behavior and it is only considered as a side effect of nonlinearity of ANNs. This view is further undermined when it was shown in \cite{Tabacof_2016} that adversarial examples are not isolated points and they form dense regions in the input space.

Next, linearity hypothesis was proposed in \cite{goodfellow2014explaining}. In contrast with the low probability pockets perspective, the linearity perspective relates the adversarial phenomenon to a side effect of linear classifiers in high dimensions. According to the linearity perspective, nonlinear classifiers like ANNs are showing the phenomenon because they are trained with algorithms that prefer linear models. Using this theory, Goodfellow et al. developed the \textit{Fast Gradient Search Method} (FGSM) which was able to easily produce adversarial examples that would also transfer to other networks.

This attack is constructed as follows. Consider the inner product of the weights $w$ and an adversarial example $\tilde{x}=x+\eta$.
\begin{equation}
\langle w,\tilde{x}\rangle=\langle w,x\rangle+\langle w,\eta\rangle
\label{eq:lfgsm}
\end{equation}
According to \eqref{eq:lfgsm}, if we choose $\eta=\mathrm{sign}(w)$, the change in the result of the inner product would be maximized, constrained by $l^\infty$ norm. According to \cite{goodfellow2014explaining}, since the sensory precision is fairly limited, the input domain is discrete. But, on a computer, we have to represent these quantities using floating-point precision numbers. If the input dimension is large enough, the accumulated error of floating-point arithmetic in the output could be considerable even if $\eta$ was actually small. Linearity means that the trained classifiers show a similar behavior and would be fooled if we choose $\eta$ to be the sign of the gradient of the loss function with respect to $x$.

Linearity perspective explains adversarial transfer as the side effect of converging to the optimal linear classifier. According to \cite{tanay2016boundary}, this theory has two predictions. First, that all linear classifiers show adversarial phenomenon and second, that the effect would get worse as we increase the dimensions of the input space. Nevertheless, Tanay and Griffin showed that both of these predictions are wrong\cite{tanay2016boundary}. They constructed a linear classifier which did not show the adversarial phenomenon and they showed that the effect of the adversaries does not increase as the dimensions of the problem increases.

The proposal of Ilyas et al. is also of interest in our discussion\cite{ilyas2019adversarial}. They explain the existence of adversarial examples by attributing them to features that are predictive, but nonrobust. A useful but nonrobust feature is a feature that is highly predictive of the true label on the empirical distribution of samples and labels, but if we add adversarial perturbations to samples, it would not be as useful anymore. That is, for a  distribution $\mathcal{D}$, adversarial perturbations $\Delta(x)$ and a feature $f$ with a positive correlation with the label $y$,
\begin{equation}
\mathbb{E}_{\mathcal{D}}[y.f(x)] > \mathbb{E}_{\mathcal{D}}[y.f(x + \Delta(x))].
\end{equation}
The authors show that these features consistently exist in standard datasets and tie the phenomenon they observed to a misalignment between the human-specified notion of robustness and the inherent geometry of data. To support their proposal, authors describe a process in which robust and nonrobust features could be extracted from the training samples and showed that removal of nonrobust features from the training samples would exchange accuracy with a decrease in the adversarial phenomenon effects.

There are other proposal and studies as well. For example Shamir et al. proposed that the phenomenon is a natural consequence of geometry of $\mathbb{R}^n$ with Hamming metric\cite{DBLP:journals/corr/abs-1901-10861}, or Tanay and Griffin proposed that the adversarial phenomenon can be reproduced by changing the angle of the decision boundary manifold and the manifold that the natural samples reside on\cite{tanay2016boundary}. These studies strongly suggest that the adversarial phenomenon is somehow influenced by the training set and the way classifiers are defined and trained. To the best of our knowledge, there is no approach to the problem in which all these different observations and ideas unite. In the following sections, we describe a new perspective on the cause of the phenomenon that can potentially clarify the reasons behind the existence of adversarial phenomenon and at the same time justify the observations made in the literature.
%------------------------------------------------

\section{Proposed notion}
\label{sec:sing}
In this section, we describe a setting in which the adversarial samples phenomenon occur. Our proposal is based on the proposition that convergence of ANNs to the true classifier is pointwise, not uniform. In the following, we discuss the proposition and describe its role in producing the adversarial examples phenomenon.

\subsection{Pointwise convergence}
\label{sec:pp}
Consider the training set $S=\{(0,-), (1,+)\}$. This training set consists of two samples from $[0,1]$ interval. Suppose that we want to find the maximum margin classifier of $S$. To do so, we need to choose a set of features first. Here, we use the Bernstein basis polynomials as features. The $n+1$ Bernstein basis polynomials of degree $n$ are defined as
\begin{equation}
b_{i ,n}(x)={\binom {n}{i }}x^{i }\left(1-x\right)^{n-i },\quad i =0,\ldots ,n.
\label{eq:bernbasis}
\end{equation}
Then, we choose support vector classifiers $F_n$ as our hypothesis space,
\begin{equation}
F_n(x;\alpha)=\alpha_0\sum_{i=0}^nb_{i,n}(0)b_{i,n}(x)+\alpha_1\sum_{i=0}^nb_{i,n}(1)b_{i,n}(x).
\label{eq:svm}
\end{equation}

It could be seen from \eqref{eq:bernbasis} that $b_{i,n}(0)=\delta_{i,0}$ and $b_{i,n}(1)=\delta_{i,n}$ where $\delta$ is the Kronecker delta function
\[
\delta_{i,j}={\begin{cases}0&{\text{if }}i\neq j,\\1&{\text{if }}i=j.\end{cases}}
\]
As a result, the maximum margin classifier $f_n(x)$ of degree $n$ would be defined as 
\begin{equation}
f_n(x)=x^n-(1-x)^n.
\label{eq:classseq}
\end{equation}
The above equation defines a sequence of maximum margin classifiers. For example, $f_1(x)=2x-1$ is the linear maximum margin classifier of $S$. It could be checked that $f_n$ is robust and does not show any adversarial regions.

To illustrate the phenomenon, we pose the same classification task on $S$, but use the first $n+1$ shifted Chebyshev basis polynomials as features instead. Unlike Bernstein basis polynomials, Chebyshev basis polynomials do not guarantee uniform convergence and reveal the error caused by pointwise convergence. Shifted Chebyshev polynomials of the first kind are defined as
\begin{equation*}
T_{i}^{*}(x)=T_{i}(2x-1).
\end{equation*}
and Chebyshev basis polynomials of the first kind are defined through the recurrence relation
\begin{align*}
T_0(x)&=1,\\
T_1(x)&=x,\\
T_{i+1}(x)&=2xT_i(x)-T_{i-1}(x).
\end{align*}

Similar to the case of Bernstein basis polynomials, the support vector classifiers $G_n$ are defined by
\begin{equation}
G_n(x;\alpha)=\alpha_0\sum_{i=0}^nT_i^*(0)T_i^*(x)+\alpha_1\sum_{i=0}^nT_i^*(1)T_i^*(x).
\label{eq:svmcheb}
\end{equation}
It is known that $T_i^*(1)=1$ and $T_i^*(0)=(-1)^i$. As a result, the maximum margin classifier $g_n(x)$ of degree $n$ would be defined as
\begin{equation}
g_n(x)=\frac{2}{n+1}\sum_{i=0}^{[\frac{n-1}{2}]}T_{2i+1}^*(x).
\label{eq:classseqcheb}
\end{equation}

\begin{figure}
    \centering
    \includegraphics[width=0.45\textwidth]{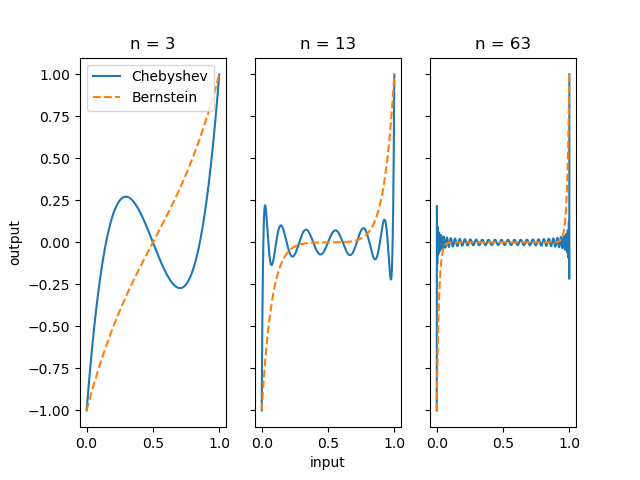}
    \caption{A comparison of the Bernstein and Chebyshev maximum margin classifiers of $S$. It could be seen that convergence of Chebyshev polynomials is not uniform.}
    \label{fig:pointwise}
\end{figure}

Figure \ref{fig:pointwise} compares a few members of $f_n$ and $g_n$. It could be seen that points exist that maximize or minimize the estimation error $g_n(x)-f_n(x)$. The result is oscillations in the output of $g_n$. Due to the sign of $g_n$ alternating between every root, predictions of $g_n$ agrees  with $f_n$ at most half of the time. For a large $n$, the points that do not agree become dense in the points that do. Nonetheless, for a small $n$, they produce measurable regions in the input space.

To put the phenomenon in learning theory terms, the hypothesis space $\mathcal{P}$ of polynomials is nonuniform learnable. Hence, the hypothesis space $\mathcal{P}_n$ of polynomials of degree $n$ has the uniform convergence property. Bernstein polynomials of degree $n$ are testament to this fact since they guarantee uniform convergence. A truncated Chebyshev polynomial on the other hand only guarantees pointwise convergence. As a result, even though as a hypothesis class it enjoys the uniform convergence property, a Chebyshev polynomial falls under the notion of consistency as a learning rule . We refer the reader to \cite{shalev2014understanding} for an overview of nonuniform learnability and consistency.

We propose that the phenomenon occurs when a universally consistent learning rule on a nonuniform learnable hypothesis class does not guarantee uniform convergence. According to the definition of consistency, sample complexity of a consistent learning rule depends on the generating distribution of the data as well as the hypothesis. As a result, there could be distributions of data that maximize or minimize the sample complexity of the learning rule for any hypothesis. To make it precise, the probability $\delta$ of the classifier making an error approaches zero, but the maximum possible error $\epsilon$ of the classifier does not vanish as fast as $\delta$ as we draw more samples from a sub-optimal training distribution.

To showcase the proposal, we keep the degree of the Chebyshev polynomial constant and instead increase the number of training points. The training points get labeled according to the nearest neighbour classifier of $S$. We construct two training sets, $E_m$ and $C_m$. $E_m$ is constructed using an equispaced grid of $m$ points, $C_m$ on the other hand use a Chebyshev grid as samples.

\begin{figure}
    \centering
    \includegraphics[width=0.45\textwidth]{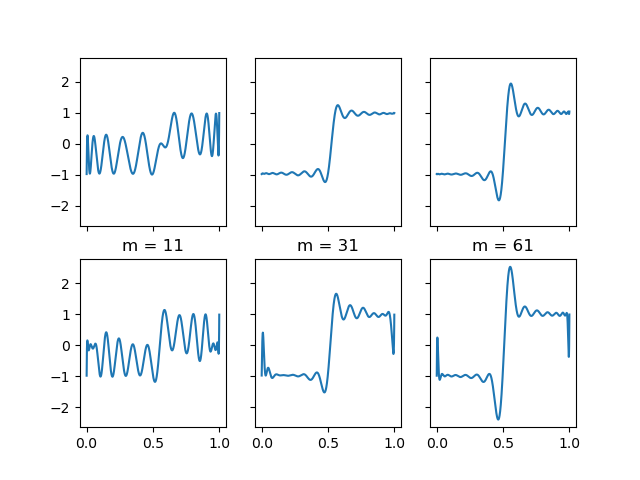}
    \caption{The trained Chebyshev maximum margin classifiers of degree 30 on $C_m$ (top row) and $E_m$ (bottom row). We expect that the predictor converge for $m>30$ on the equispaced grid. In practice though, it does not converge even for relatively large $m$.}
    \label{fig:chebgrid}
\end{figure}

The Chebyshev maximum margin classifier of degree 30 of $E_m$ and $C_m$ is depicted in Figure \ref{fig:chebgrid} for a few choices of $m$. It could be seen in the figure that in case of $E_m$, uniform convergence does not occur even for relatively large $m$. In contrast, classifiers trained on $C_m$ converge in the expected number of samples. As a matter of fact, Chebyshev polynomials guarantee uniform convergence on a Chebyshev grid, and hence, do not suffer from the phenomenon.

\subsection{Optimal points of a hypothesis class}
In the previous section, we demonstrated that Chebyshev grid optimizes the sample complexity of Chebyshev polynomials. These points seem to be reflecting the position of the critical points of the estimation error $g_n(x)-f_n(x)$. Thus, we define the optimal training points of a hypothesis class in a way that reflects this observation.

\textbf{Optimal training points:} Let $\mathcal{X}$ be a domain set, let $\mathcal{H}$ be a hypothesis class and let $A$ be a universally consistent learning rule with respect to $\mathcal{H}$. For every $X\subset\mathcal{X}$ and every $h\in\mathcal{H}$, let 
\[
\hat{h}_X=A(\{(x,h(x))\,|\,x \in X\}).
\]
The optimal training points of $A$ is a solution to the following problem plus the boundary $\partial\mathcal{X}$ of $\mathcal{X}$,
\begin{argmini}
{X}{\int_{\mathcal{H}}\sum_{x\in X}\|\nabla(\hat{h}_X(x)- h(x))\|\,dh}{\label{opt:optim}}{}
\addConstraint{X}{\mathrm{\,is\,feasible}}{}
\end{argmini}

The intuition behind the definition of the optimal points is that by minimizing the empirical risk on the critical points of the error function, we are effectively minimizing the maximum norm of the true error.

Computing the objective of \eqref{opt:optim} is not tractable for any practical purpose. However, steps could be taken to calculate an approximation. First, we can utilize the loss function $L$ as a surrogate for the objective of \eqref{opt:optim}. We know that the loss function is proportional to the magnitude of the error function,
\[
   L(h(x),h(x)) \propto |\hat{h}_X(x)- h(x)|.
\]
Hence, points that maximize $L(h(x),h(x))$ would be critical points of $\hat{h}_X(x)- h(x)$ and consequently would minimize $\|\nabla(\hat{h}_X(x)- h(x))\|$. Of course, this is only a heuristic and critical points of the error function could exist that are local minimas of $|\hat{h}_X(x)- h(x)|$. Second, we approximate the integral in \eqref{opt:optim} by restricting the hypothesis class. We call the solution to this approximation the hard training points of the restricted class.

\textbf{Hard training points of $H$:} The hard training points of $A$ with respect to a distribution $H$ on $\mathcal{H}$ is a solution to the following problem,
\begin{argmaxi}
{X}{\mathbb{E}_H[\sum_{x\in X}L(h(x),h(x))]}{\label{opt:best}}{}
\addConstraint{X}{\mathrm{\,is\,feasible}}{}
\end{argmaxi}

Looking at \eqref{opt:best}, we can see that it is similar to the program that results in adversarial examples. Indeed, if we reduce $H$ to a specific $h$, we would find an approximation to \eqref{opt:optim} that closely resembles the objective of evasion attacks.

\textbf{Adversarial training points of $h$:} The adversarial training points of $A$ with respect to $h\in\mathcal{H}$ is a solution to the following problem,
\begin{argmaxi}
{X}{\sum_{x\in X}L(h(x),h(x))}{\label{opt:adv}}{}
\addConstraint{X}{\mathrm{\,is\,feasible}}{}
\end{argmaxi}

There are two differences between the proposed definition of the adversarial training points of $h$ and the standard definition used in adversarial training. First, our definition does not use the true labels in computing the loss. While the standard definition may lead to stronger adversaries, our proposal makes the independence of the phenomenon from the true predictor explicit. Second, Our definition does not require the points to be close to natural samples.

\section{Experiments}
\label{sec:exp}
In section \ref{sec:sing}, we introduced a framework in which adversarial examples are produced. Here, we back the propositions by providing empirical evidence in their support. In our first experiment, we show that a Chebyshev grid does optimize \eqref{opt:best}. Then, we demonstrate how \eqref{opt:best} could be used to adversarially train a MLP.

\subsection{Hard training points of Chebyshev polynomials}
To exhibit that the definition of hard training points is consistent with the observations made in section \ref{sec:sing} about Chebyshev polynomials, we now derive and visualize the objective of \eqref{opt:best} for a special family of polynomials in a maximum margin classification setting.

It is known that the hinge loss function is the loss of the maximum margin classifier. Thus, we are looking for hard training points of a shifted Chebyshev polynomial of degree $n$ with respect to the hinge loss function.
\begin{argmaxi}
{X}{\mathbb{E}_H[\sum_{x\in X}\max(0,1-h(x)^2)]}{\label{opt:hinge}}{}
\addConstraint{0 \leq x_1<...<x_m \leq 1}{}{}
\end{argmaxi}

The $\max$ operator in the objective of \eqref{opt:hinge} would make the objective hard to manipulate. As a result, we restrict ourselves to a family of polynomials with range in the $[-1,1]$ interval. Shifted Chebyshev basis polynomials are examples of such polynomials. Using shifted Chebyshev basis polynomials as $H$, the problem will change to a form that is easier to analyse and compute.
\begin{argmaxi}
{X}{\sum_{x\in X}\frac{1}{n+1}\sum_{i=0}^n 1-T_i^*(x)^2}{\label{opt:shcheb}}{}
\addConstraint{0 \leq x_1<...<x_m \leq 1}{}{}
\end{argmaxi}

From \eqref{opt:shcheb} it could be deduced that the local minimas of $\sum_{i=0}^n T_i^*(x)^2$ solve \eqref{opt:shcheb}. Next, it could be argued that the optimal solution is close to the roots of $T_n^*(x)$, which are exactly the points of a Chebyshev grid.

\begin{figure}
    \centering
    \includegraphics[width=0.45\textwidth]{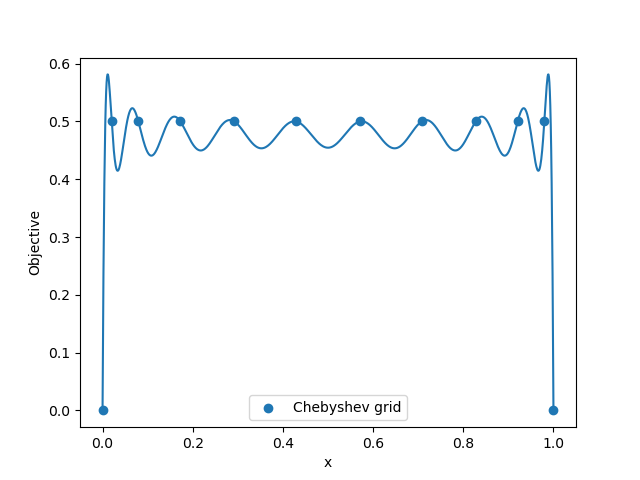}
    \caption{The expected value of loss on $H$ in \eqref{opt:shcheb} for $n=10$. The points mark the value of the expectation at each node of a Chebyshev grid.}
    \label{fig:optgrid}
\end{figure}

Figure \ref{fig:optgrid} compares the position of Chebyshev nodes with $\frac{1}{n+1}\sum_{i=0}^n 1-T_i^*(x)^2$. It could be checked in the figure that Chebyshev nodes are very close to the optimal points. The two nodes on the two end of $[0,1]$ are also critical points of the error function because they belong to the boundary of the interval. In our experiments, removal of the two boundary points has a considerable negative effect on the trained classifier. This shows that we should consider the boundary $\partial\mathcal{X}$ of the domain set as well.

\subsection{Optimal training points in MLPs}
We showed that the proposed notion can explain the phenomenon in case of polynomials. Here, we analyze MLPs through our framework. To this end, we sample a random MLP classifier and visualize the adversarial and hard training points objectives with respect to that network.

The domain set of the network is $[-1,1]^2$. The dimensions of the network are $2\times100\times100\times100\times5$. The weights of the network are sampled independently from a standard normal distribution. The activation function of the hidden layers is $\tanh$ and the output of the network is activated through softmax. We have chosen the cross entropy loss as our surrogate for the error function.

The objective of \eqref{opt:adv} is straightforward to compute. To compute \eqref{opt:best}, we need to choose a distribution over the hypothesis class of MLPs. Following the case of Chebyshev polynomials, we use each neuron in the feature layer of the MLP as the hypothesis set. In other words, we create a copy of the network, but replace the last layer with an identity transform and compute \eqref{opt:adv} for the new network. For this particular example, the size of the resulting network would be $2\times100\times100\times100\times100$.

\begin{figure}
    \centering
    \includegraphics[width=0.45\textwidth]{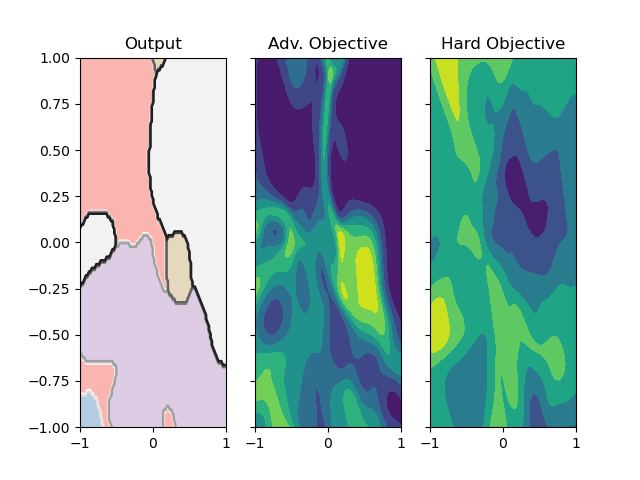}
    \caption{A comparison between the output, the adversarial objective and hard objective of a random network. The adversarial objective is representative of the decision boundary. On the other hand, the hard objective suggests a very different distribution of optimal points.}
    \label{fig:land}
\end{figure}

Figure \ref{fig:land} shows the results. By comparing the output of the original network with the adversarial objective of the network, we can see that the adversarial objective is sensitive to the position of the decision boundary. But, we can also identify some regions of the objective that get lighter or darker without respecting the position of the decision boundary. According to our proposal, if we sample the training points proportional to the adversarial objective, the sample complexity of finding this exact hypothesis would be minimized.

Contrasting the adversarial objective and the hard objective in Figure \ref{fig:land} reveals that a hypothesis agnostic set of points would not be as helpful as for the case of polynomials. In other words, while we could reach a definitive set of optimal points for polynomials, the optimal points of MLPs are not unique even up to the layers of a single network. As a result, even though the proposed notion can explain the existence and abundance of adversarial examples in MLPs, it cannot further explain their transfer between different architectures of MLPs. Specially, we could marginalize out the hypothesis space from the sample complexity in polynomials, but the same could not be said about MLPs.

\subsection{Adversarial training of a MLP}
In this section, we show that sampling according to the hard objective described in the previous experiment does indeed optimize the sample complexity of the robust hypothesis. To do so, we adversarially train a MLP on a subset of MNIST using the hard training points of the feature layer. Next, we train a MLP with the same architecture but this time use the whole of MNIST as the training set with no adversarial training. Then, we compare the performance of the networks in both adversarial and nonadversarial settings. To show that the results are consistent across multiple real-world datasets, we conduct the same experiments for Fashion-MNIST\cite{xiao2017/online} and report the results.

Here we use $784\times512\times512\times10$ networks with ReLU activation. The robust network is trained using a third of the full datasets. We have trained each network for 10 epochs. In each step of the adversarial training, we find the hard training points close to the batch with regards to the weights of the feature layer in that step. To be more precise, we use the current batch as the initial points and take a few gradient ascend steps in the direction of \eqref{opt:best}. Then, we train the network on this new batch of samples as well.

\begin{figure}
    \centering
    \begin{subfigure}[b]{0.45\textwidth}
     \centering
     \includegraphics[width=\textwidth]{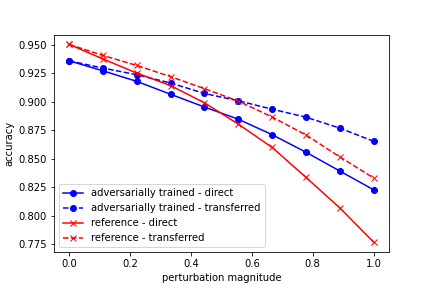}
     \caption{MNIST}
    \end{subfigure}
    \hfill
    \begin{subfigure}[b]{0.45\textwidth}
     \centering
     \includegraphics[width=\textwidth]{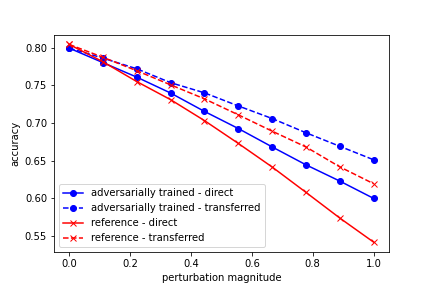}
     \caption{Fashion-MNIST}
    \end{subfigure}
    \caption{The accuracy of MLPs compared to the magnitude of the adversarial perturbation. The data shows that the adversarial examples of the adversarially trained network does transfer to the reference network, but not vice versa.}
    \label{fig:attack}
\end{figure}

Figure \ref{fig:attack} shows the accuracy of all networks on the test set as a function of the magnitude of the adversarial perturbation. A $L^2$ normalized one-step gradient based attack is used for this experiment.

It could be seen that the adversarially trained network gets a comparable performance with the reference network on the natural samples set even though it is trained on a third of the samples. The adversarially trained network is also more robust compared to the reference network if we compare their performance in the direct and transfer attack settings. The results show that adversarial training using \eqref{opt:best} would make the trained MLP more robust against adversarial attacks by attacking random hypothesises that use the same feature layer instead of attacking the hypothesis itself.

\section{Conclusion and future work}
\label{sec:diss}
In this paper, we introduced a framework to explain the adversarial examples phenomenon. We defined the adversarial examples as the critical points of the error function. We showed that this principle can fully explain adversarial examples existence, training and transfer for pointwise converging polynomials.

Unfortunately, this approach does not enjoy the same success in case of MLPs. We could demonstrate that MLPs do follow the principle in case of existence and training, but the same principle proved to be insufficient in explaining transferable adversarial examples. Nevertheless, the results suggest that the phenomenon would disappear if the learning process guarantee uniform convergence of MLPs.

With regards to other proposals in literature, our definition is more aligned with the low probability pockets perspective. We constructed the first instance of a classifier with adversarial examples that are dense in the input domain. While the low probability pockets perspective relates the phenomenon to highly nonlinear nature of ANNs, we have shown that the phenomenon does occur for low degree polynomials as well, which would produce adversarial regions in turn.

With respect to the linearity hypothesis, our analysis does not confirm the view that the phenomenon is a result of representing the input by floating-point numbers. We also introduced an alternative explanation for the success of gradient based attacks that maximize the training loss function. While our proposal does not confirm that adversarial examples transfer is due to converging to the optimal linear classifier, we share the idea that the transfer is rooted in properties of the optimal classifier.

The proposed notion is similar to the nonrobust features perspective in that we also relate the phenomenon to a form of optimal training points. Our proposal can accommodate a notion of pointwise converging features similar to nonrobust features as well. On the other hand, we do not support the idea that adversarial examples transfer in MLPs is caused by patterns in the input that are imperceptible to humans. In case of MLPs, we did not observe any hypothesis independent relation between the distribution of the optimal points and the domain set in our experiments.

Our framework is heavily inspired by analogous concepts in approximation theory. For now, we have analysed both the training points and the hypothesis class by introducing a definition for optimal training points and revealing the role of convergence properties of the hypothesis class in the phenomenon. If approximation theory is any indicator of the path forward, our next step should be to analyse the classifier independently from the training points. In other words, we guess that there is something special about generalizing decision boundaries that is causing the transfer of the adversarial examples in MLPs. In future work, we would focus on finding extensions to the framework that would help explain the transfer of adversarial examples in MLPs.

\bibliographystyle{IEEEtran}
\bibliography{IEEEabrv,example_paper}

\end{document}